\documentclass[fleqn,twoside]{article}
\usepackage{espcrc2}
\usepackage{graphicx}

\newcommand{\AmS}{{\protect\the\textfont2 A\kern-.1667em\lower.5ex\hbox{M}\kern-.125emS}}

\def\be{\begin{equation}}
\def\ee{\end{equation}}

\newcommand{\ba}{\begin{eqnarray}}
\newcommand{\ea}{\end{eqnarray}}
\newcommand{\x}{\mbox{$\vec{x}$}}
\def\ket#1{\vert #1 \rangle}
\def\bra#1{\langle #1 \vert}
\def\bracket#1#2{\langle #1 \vert #2 \rangle}

\newcommand{\Schrodinger}{Schr\"odinger\ }

\newcommand{\hs}{\hskip - 2pt}
\title{\noindent\hfill\hbox to 1.5in{\rm  } \vskip 1pt \noindent\hfill\hbox
to 1.5in{\rm SLAC-PUB-13759 \hfill  } \vskip 1pt
\vskip 10pt
Strange Bedfellows:  Quantum Mechanics and Data Mining}
\author{Marvin Weinstein\address[SLAC]{SLAC National Accelerator Laboratory,\\
        Stanford, CA, USA}%
        \thanks{This work was supported by the U.~S.~DOE, Contract No.~DE-AC02-76SF00515.}}

\begin{document}

\begin{abstract}
Last year, in 2008, I gave a talk titled {\it Quantum Calisthenics\/}.  This
year I am going to tell you about how the work I described then has spun off
into a most unlikely direction.  What I am going to talk about is how one
maps the problem of finding clusters in a given data set into a problem in
quantum mechanics.  I will then use the tricks I described to let quantum
evolution lets the clusters come together on their own.
\vspace{1pc}
\end{abstract}

\maketitle

\section{What Is This About ?}

Since I am talking to you at Light Cone 2009, it is fun for me
to point out that a good deal of the material I will present is a
direct spin-off of material I presented last year in my talk,
{\it Quantum Calisthenics\/}.  What is new (and surprising to me)
is that the ideas for solving problems in quantum mechanics would prove
extremely useful in dealing with data-mining, a problem in computer
science that, at first glance, does not appear to have anything to
do with quantum physics, or for that matter, with any kind of physics.
In the next few minutes hope to convince you that despite
appearances data-mining and quantum physics are a match made in heaven.

\subsection{What Is The Problem And How Does It Affect You ?}

As one wag said, the problem of data-mining can be summarized by,
"If a supermarket customer buys formula and diapers,
how likely are they to buy beer?".  Clearly, if it is possible to
give a good answer to this question, a supermarket manager
can decide if it is better to put the beer next to the diapers, or
to put it at the other end of the store to encourage impulse buying.

As amusing as this simple summary may be, there is a bit more to the
story. For example, other businesses that engage in data-mining are
Amazon and NETFLIX.  Both companies have large data bases containing
the information about each customer's previous purchases.  Their
goal in mining this data is to suggest what the
next book a given customer might like to purchase, or movie she
might like to rent.  They base their suggestion on what
they believe customers {\it like her\/} have purchased or rented.
The key question is how do they determine who is {\it a customer
like her ?}  In general the examination of unstructured data to find
{\it clusters\/} that are similar in some
way is a key element of data-mining.  The search for these
structures in a data-set is imaginatively referred to as
{\it clustering\/}.

While the use of clustering by Amazon and NETFLIX may improve the
shopping experience, data-mining affects all of us in ways that
impact out existence at a much more important level.  This happens
when banks and insurance companies apply it to the problem of {\it
scoring\/}. Basically, the idea is to take all of the data they have
about people and identify clusters of similar people so as to
predict how likely it is that an applicant will default on a
mortgage, or loan, or how likely they are to be robbed, die, etc..
To do this they take their and try, in various
ways, to find clusters of people that they can define as similar.
For better or worse we are all affected by these practices.

Finally, to give one of a myriad of examples that have nothing to do
with questions of business, let us consider the case of gene chips as
a tool for medical diagnosis and treatment planning.  Gene chips are
a technology for measuring which genes are being under or
over-expressed by a cell.  The idea is that understanding
the different patterns of gene expression and how they relate to
specific diseases and drug resistance will allow a doctor to both
diagnose a disease and prescribe the correct drug for treating it.
Later I will talk about a specific example of this sort of study
for the case of patients with Type A or Type B Leukemia.

Summarizing we can say that many fields - physics, biology, medicine
intelligence and homeland security, finance, insurance, and diplomacy -
all collect large databases of information.  Sorting through such data
and searching for previously unknown interesting and relevant structures
is referred to as data-mining.  Clearly from the breadth of
topics one might guess that no precise definition of data-mining exists.
However, although we can't make the definition of what we wish to do
totally specific, it is intuitively clear that it is a useful notion.
In what follows I will be describing a clustering technique that my
colleague David Horn and I call DQC (Dynamic Quantum Clustering).
I hope that the specific examples that I discuss will give you a
flavor for the general problem.  A more complete exposition of this
material is given in Ref.\cite{MandH}.

\section{The Data Miner's Lament: The Curse of High Dimension}

The general problem is we are given a database with a large number of
entries.  Associated with each entry is a record that contains all
of the information that has been collected about that entry, each
such item of information is called a {\it feature\/}.  Clearly the
number of features and the type of information stored in each
feature varies from database to database.  As an example, let's talk
about NETFLIX. Assume that they have a library of 10,000 films that
can be rented.  Then each entry in the database might correspond to
a customer and associated to each customer would be 10,000 features
(in numerical format) that indicate which films the customer has
rented and, if available, his/her rating of each of those films.
The number of features is called {\it the dimension of the problem\/}.
Thus, we see that finding clusters of people with similar tastes in
the NETFLIX problem involves searching for clusters in a space of
very high dimension.  It should not surprise you that this is
computationally difficult.

Because it is so difficult to search for clusters in spaces of high
dimension, people try to find a way to reduce the dimension of the
problem without, too badly, compromising the information.  The tool
most commonly used to accomplish this is SVD or Singular Value
Decomposition.  Clearly, given the time available to me I can't
go into the use of SVD too deeply, but I would like to take a
few moments to give you a feeling for what it does and doesn't
do.

\subsection{Singular Value Decomposition: The Swiss Army Knife of Data Mining}

SVD, or Singular Value Decomposition, is the statement that any $n
\times m$-matrix can be brought to an almost diagonal form.  The
first time I was introduced to this theorem many years ago I
couldn't believe it was really good for anything.  That just goes to
show what I knew. In fact, nowadays, SVD has crept into all sorts
of areas of physics and data-mining.  It is extensively used in
solid state physics computations based on the DMRG (or density
matrix renormalization group) method, in image compression, in
data-mining to do - among other things - dimensional reduction,
and a host of other places.  There is much I could say to try and
expose you to its many applications, but time being what it is I
will limit myself to giving a brief introduction to the SVD
decomposition and an example of how we will use it.

Basically SVD is useful to us because all of the data we will mine
will be presented in the form of an $n \times m$-matrix of
numerical values.  Each row of this matrix will correspond to
the entries in the data base (i.e., customers, events in a physics
experiment, the specific cell culture corresponding to a given
gene-chip map, etc.).  The columns of the matrix represent the
features that have been measured for each entry in the database.
The theorem establishing the SVD decomposition says that for
any $n \times m$-matrix $M$, there exists a unitary
$n \times n$-matrix $U$, an $n \times m$-matrix $S$ and a unitary
$m \times m$-matrix $V$, such that
\begin{equation}
    M = U\,S\,V^\dag .
\end{equation}
where the matrix $S$ only has non-vanishing entries along the
diagonal.  Consider the explicit example:
{\scriptsize
\begin{eqnarray}
\left(
      \begin{array}{cc}
        \hs\hs 8.3 \hs & \hs 5.8\hs\hs \\
        \hs\hs -4.5\hs&\hs-4.3\hs\hs \\
        \hs\hs 6.8\hs&\hs-8.5\hs\hs \\
      \end{array}
    \right) \hs\hs\hs\hs&=&\hs\hs\hs\hs
    \left(
      \begin{array}{ccc}
        \hs-0.85\hs& \hs0.077\hs &\hs-0.522\hs\hs\\
        \hs\hs0.51\hs&\hs-0.133\hs&\hs-0.85\hs\hs\\
        \hs\hs-0.135\hs&\hs-0.988\hs&\hs0.074\hs\hs \\
      \end{array}
    \right)\cdot \hfill \nonumber\\
    &&\hskip -15pt \left(
      \begin{array}{cc}
        \hs\hs 11.875\hs &\hs 0\hs\hs \\
        \hs\hs 0 \hs&\hs 10.907\hs\hs \\
        \hs\hs 0 \hs&\hs 0 \hs\hs\\
      \end{array}
    \right)\cdot
    \left(
      \begin{array}{cc}
       \hs\hs -0.864 \hs&\hs -0.503\hs\hs \\
       \hs\hs -0.503 \hs&\hs 0.864 \hs\hs\\
      \end{array}
    \right),
\end{eqnarray}}
\unskip where $M$ is the matrix on the left hand side of the equation and
$U$, $S$ and $V^\dag$ are the three matrices on the right hand side
of the equation.

Now, to get a feeling for what is decomposition is doing for us, let
us think of the rows of $M$ as the coordinates of three different points in two
dimensions.  Clearly these three vectors cannot all be linearly independent as
there can only be two linearly independent vectors in $2$-d.
The SVD decomposition gives us a coordinate system for plotting these
points that is adapted to the way the data is spread out.  To be
more specific we note that the two rows of $V^\dag$ are two ortho-normal vectors
that span the space containing the points.  From the formula for $M$ we
see that each of the original points can be written as linear combination
of the two rows of $V^{\dag}$.  The coefficient of each of these two basis
vectors is given by the entries in the first two rows of $U$ times the
corresponding elements of $S$.  From this we see that the rows of $U$ serve
as coordinates for each of the data points, where the data has been mapped
onto the unit sphere (since each row of $U$ is a unit vector).
If one wishes to {\it dimensionally reduce\/} the clustering problem
one can use a smaller number of columns of the $U$-matrix to label the
points.  In this case we typically rescale each row to have unit length.
In all of the examples that follow we have done this step before starting
the DQC procedure.

Another use of SVD is to reduce the amount of data used to describe a picture
as either a way of filtering out noise, or as a first step towards image
recognition.  To see how this works assume that a picture is given by
a matrix, $M$, of pixels.  Another way to write the SVD decomposition of
this matrix is
\be
    M_{ij} = \sum_{\nu = 1}^{N_{max}} S_{\nu \nu}\, t^{\nu}_{i,j}
\ee
where the numbers $\lambda_{\nu}$ run over the non-zero numbers on the
diagonal of $S$ (in decreasing order) up to a maximum, $N_{max}$,
and the matrices $t^{\nu}_{ij}$ are given by
\be
    t^{\nu}_{i,j} = U_{i \nu}\, V^{\dag}_{\nu j}.
\ee
This is a useful way of looking at things because the
diagonal entries in $S$ tend to decrease rapidly.  Thus,
since $U$ and $V^{\dag}$ (and thus $V$) are unitary matrices, we see
that
\be
    \sum_{i,j} M^2_{ij} = \sum_{\nu} S_{\nu \nu}^2 .
\ee
If we approximate the original picture by a sum over the $N$
largest entries in $S$; i.e.,
\be
    M^{approx}_{ij} = \sum_{\nu = 1}^{N} S_{\nu \nu}\,t^{\nu}_{ij}
\ee
then it follows that
\be
   \sum_{i,j}(M - M^{approx})^2_{ij} = \sum_{\nu = N+1}^{N_max} S_{\nu \nu}^2
\ee
and so, we see that the sum of the squares of all of the errors in the
pixels is bounded by the sum of the square of the eigenvalues $S_{\nu \nu}$
that weren't included in the approximation.

\subsection{What About Clustering ?}

To this point I have only been talking about using the SVD
decomposition to plot the data in a coordinate system that is
adapted to the natural structure of the data. Thus the first axis is
the one in which the data had the largest variance; i.e. had the
biggest difference between the largest and smallest coordinate
values, the second axis is the direction in which the data has the
next largest variance, and so on.  While this step is useful and
dimensionally reducing the data by keeping only the coordinates
corresponding to the first few directions can lead to plots that in
some loose sense maximally separate data belonging to different
clusters, in general this is only a first step.

In the generic situation, especially for data that cannot usefully
be reduced to two or three dimensions, clusters cannot be easily
identified.  The situation is even more complicated if the data does
not group into separated globular clusters, but instead is
concentrated on some other geometrical structure, say for example, a
ring.  In this case SVD followed by strong dimensional reduction can
incorrectly lead one to conclude there are clusters where none
exist.  This is where DQC comes into play. DQC allows one to explore
more dimensions of the original data and it can clearly indicate the
presence of higher dimensional geometries when they exist.

\section{DQC: What Is Dynamic Quantum Clustering?}

In Dynamic Quantum Clustering, DQC, we map the problem of finding
clusters into a problem in quantum mechanics and then use the
techniques of quantum mechanics to allow the clusters, or over
dense regions in the data, to reveal themselves.

\subsection{The Parzen Window Estimator}

An old approach in clustering was the so-called Parzen Window
Estimator. The crux of this idea was that given $n$-dimensional
data, denoted by points $\vec{x_i}$, one could construct a function
on the $n$-dimensional Euclidean space as follows:
\be
    \psi(\vec{x}) = \sum_i
e^{-{1 \over 2 \sigma^2}(\vec{x} - \vec{x_i})\cdot (\vec{x} - \vec{x}_i)}
\label{Parzen}
\ee
In other words, the Parzen estimator is simply a sum of $n$-dimensional
Gaussians with a common width determined by the parameter $\sigma$.
The notion was that in regions where the data was denser, the Parzen estimator
would have a relative maximum.  The hope was that clusters could be
identified by finding these maxima and selecting a region about each
one, so that the points in that region would be said to belong to a
cluster.  There were several problems with the Parzen estimator approach.
First, finding the maxima of a complicated function in an $n$-dimensional\
space is computationally time consuming.  Second, the structure of the
function proved to depend sensitively on the choice of $\sigma$.  For
a value of $\sigma$ that is too small, one obtains a function with many
local maxima and very small clusters.  If $\sigma$ is to large the
maxima all coalesce and no distinct clusters can be found.

\subsection{Quantum Clustering}

Quantum clustering, (QC), begins with the construction of a Parzen
estimator\ref{Parzen}, but instead of using the Parzen estimator to
find clusters directly, we use it to construct a potential function
whose minima are related to the clusters.  The construction of this
function was first introduced in the paper of Horn and
Gottlieb\cite{qc1} where it was defined by the condition that it is
the function $V(\vec{x})$ for which the Parzen estimator satisfies
the $n$-dimensional time-independent \Schrodinger\ equation
\be
\label{sch}
    H\psi = \left(-{\sigma^2 \over 2}\vec{\nabla} ^2 + V({\vec{x}})\right)\psi = 0 .
\ee
That such a $V(\vec{x})$ always exists is clear, all one has to do
is to solve equation \ref{sch} for $V(\vec{x})$; i.e.,
\be
\label{Vofx}
  V(\vec{x}) =
  -{\sigma^2 \over 2\, \psi(\vec{x})}\,\vec{\nabla} ^2 \,\psi(\vec{x}) .
\ee
The intuition behind this prescription is based upon the fact that in
the quantum problem local maxima in the ground state wave-function
correspond to local minima in the potential; moreover, such minima are
likely to be a good deal sharper and deeper than the corresponding
maxima of the wave-function, due to the fact that the kinetic term,
$-\vec{\nabla}^2$, causes the wave-function to be more spread out
than the features of the potential would suggest.  A sample of such
a potential function will be shown in the next section when I
discuss the example of Ripley's Crab Data.

In the orginal quantum clustering approach\cite{qc1} points lying in
the basin of attraction of particular minimum were identified as a
single {\it cluster}. One way of determining which minimum was
closest to a given point was to classically {\it roll the points
downhill\/} using the gradient descent method.  The disadvantages to
this method are: first, it can be computationally very expensive to
carry out when there a large number of points and a large number of
features involved; second, there can be many instances where a slightly
too small value for $\sigma$ can lead to a potential with several
nearby shallow minima contained in a single larger minimum. When this
happens the classical prescription will lead to several small
clusters and not show the existence of the larger cluster.  As we
will see, by appropriately choosing one of the two parameters that
control the DQC evolution, this sensitivity upon the choice of
$\sigma$ can be greatly reduced and the structure of such nearby
minima can be explored even in problems with a large number
of features.

\subsection{Ripley's Crab Data}

As a simple example of the construction of the potential associated
with a data set let us consider the data on the morphology of rock
crabs that was used in the book on {\it Pattern Recognition and Neural
Networks\/} by R.D.Ripley\cite{Ripley}.  The problem in this
exercise is to see how well one can classify the members of a
collection of 200 crabs; the collection contains 100 crabs of
two different species and each set of 100 is further subdivided
into 50 crabs of each sex.  The only information one has consists
of five measurements of carapace size and claw size.  In the picture
I have shown I have plotted a potential in two-dimensions.  The
colored dots correspond to real classes.  To construct this
potential we did an SVD decomposition of the data-matrix and then
used the second and third principal component to construct the
potential.  As you can see from the picture the potential has
four local minima and, except for a few outliers, the different
species and sexes are identified by the valley in which they lie.

\begin{figure}[h!]
 \hbox to \hsize{\hss
  \includegraphics[width=2.5in]{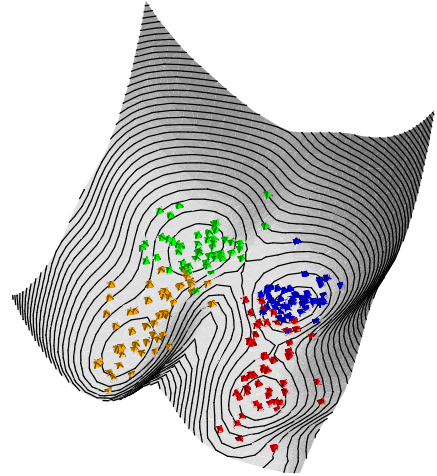}
  \hss}
  \caption{This is a plot of the quantum potential function for
  the two-dimensional problem of the Ripley's Crab Data where
  the coordinates of the data points are chosen to be given by
  the second and third principal components.  The four known
  classes of data points are shown in different colors and
  are placed upon the potential surface at their original locations.
  }
\label{twodpot}
\end{figure}

\subsection{DQC: Dynamic Quantum Clustering}

DQC begins by constructing the same potential function used
in quantum clustering, it differs in how one handles the
problem of how to identify data points with local minima of
the function in $N$-dimensions.  We simply exploit
the fact that from the outset we are dealing with a
quantum problem and a well defined Hamiltonian.  To roll
the different data points down hill we simply evolve each
Gaussian wave-function using the time development operator
\be
    U(t) = e^{-i\,H\,t},
\ee
so that the evolution of any state $\psi_i(\vec{x})$ is
defined to be
\be
    \psi(\vec{x},t) = e^{-i\,H\,t}\,\psi(\vec{x}).
\ee

This time evolved state is the solution to the time-dependent
\Schrodinger\ equation
\ba
\label{Schdeqn}
i {\partial \psi_i(\vec{x},t) \over
\partial t}
&=& H \psi_i(\vec{x},t) \nonumber\\
&=& \left(-{\nabla^2 \over 2m}
+ V(\vec{x})\right)\psi_i(\vec{x},t) ,
\ea
The important feature of quantum dynamics, which makes the evolution
so useful in the clustering problem, is that according to Ehrenfest's
theorem, the time-dependent expectation value
\be
 \bra{\psi(t)}\,\x\, \ket{\psi(t)} = \int\,d\x \, \psi^\ast_i(\vec{x},t)\,
\x\,\psi_i(\vec{x},t) ,
\ee
satisfies the equation,
\ba
   { d^2  \langle\,\vec{x}(t) \rangle \over dt^2} &=&
   - {1 \over m}  \int d\vec{x}
 \psi^\ast_i(\vec{x},t)\,\vec{\nabla}V(\vec{x}) \,\psi_i(\vec{x},t) \\
 &=&- {1 \over m}\, \bra{\psi(t)}\,\vec{\nabla}V(\vec{x})\,\ket{\psi(t)} .
\ea
If $\psi_i(\x)$ is a narrow Gaussian, this is equivalent
to saying that the center of each wave-function rolls towards
the nearest minimum of the potential according to the classical
Newton's law of motion.  This means we can explore the relation of
this data point to the minima of $V(\vec{x})$ by following the
time-dependent trajectory $\langle\,\vec{x}_i(t) \,\rangle =
\bra{\psi_i(t)}\,\vec{x}\,\ket{\psi_i(t)}$.
Clearly, given Ehrenfest's theorem, we expect to see any points
located in, or near, the same local minimum of $V(\vec{x})$ to
oscillate about that minimum, coming together and moving apart. In
our numerical solutions we generate animations which display
this dynamics for a finite time.
This allows us to visually trace the clustering of points associated with
each one of the potential minima.

At first glance this approach would seem to be a step backwards
since we have replace solving ordinary classical differential
equations by the problem of solving more complicated partial differential equations,
but this is incorrect.  The trick is to think
of the problem reduced to the subspace of the full Hilbert
space spanned by the original data points.  While this is not
completely equivalent to the original problem, for our purposes
the accuracy of the approximation is good enough to capture
the same clusters.

Clearly, these states
are not orthogonal to one another, however we can choose an orthonormal
set of states that span the subspace by computing the matrix
\be
    N_{ij} = \bracket{\psi_i(\vec{x})}{\psi_j(\vec{x})}
\ee
and then computing its eigenvectors.  The eigenvectors corresponding
to non-vanishing eigenvalues form a set of orthogonal linear
combinations of the original states and they are normalized by
dividing each eigenstate by the inverse square root of its
eigenvalue.  The fact that the initial states are all Gaussians
makes computing $N_{ij}$ analytically a trivial exercise.
Furthermore, the Gaussian nature of the original states also
makes it simple to calculate the matrix elements of the Hamiltonian
and the position operators $X_k$ (these operators act on our
wavefunctions by multiplying them by the $x_k$ coordinate).
\be
    H_{ij} = \bra{\psi_i(\vec{x})}\,H\,\ket{\psi_j(\vec{x})},
\ee
where $H$ is defined to be
\ba
    H &=& {\vec{p}^2 \over 2\,m} + V(\vec{x}) \nonumber\\
      &=& -{\vec{\nabla}^2 \over 2\,m} + V(\vec{x}).
\ea
and
\be
   X_{ij} = \bra{\psi_i(\vec{x})}\,X_k\,\ket{\psi_j(\vec{x})}.
\ee
(Note: I have introduced the parameter $m$ into the Hamiltonian
used to evolve the states.  This is different from the $\sigma$
used to construct the potential.  I do this so that by
choosing $m < 1$ I can increase the effect of quantum tunneling
and decrease the sensitivity by causing points in nearby
minima to merge.)

Given the analytical expressions for these matrix elements
between the Gaussian states it is a simple matter to compute
the same operators in the orthonormal basis defined by $N_{ij}$ and
to then exponentiate the Hamiltonian in this basis.  In this way
the apparently difficult problem of solving the time-dependent
\Schrodinger\ equation is reduced to the computation of
simple closed form expressions followed by numerical evolution
in the truncated Hilbert space.  This trick reduces the problem
to dealing with matrices whose size is determined by the number
of data points and not the dimension of the data-set (i.e., the number
of features associated with each data point).  Of course, when
there are a large number of data-points this might seem to be
an intractable problem however, fortunately, there is a
simple trick for dealing with that situation too.

Applying these ideas to Ripley's crab data, reduced to the
first three principle components, is very instructive.
(The evolution of the full five-dimensional data-set shows
exactly the same behavior but is more complicated to plot.)
In the following pictures I show two steps in the time evolution
of the initial distribution.  The first picture is a snapshot
of the original data before evolution has begun.  In the next
picture we see what happens after a short time. As expected
the points begin to approach one another.  The last picture
shows what happens if we stop the points where they are
and restart DQC using the new points.  The clustering is
quite easy to see and it is easy to understand why the
data is clustering as it is.
\begin{figure}[h!]
 \vbox{\hbox to \hsize{\hss
  \includegraphics[width=1.5in]{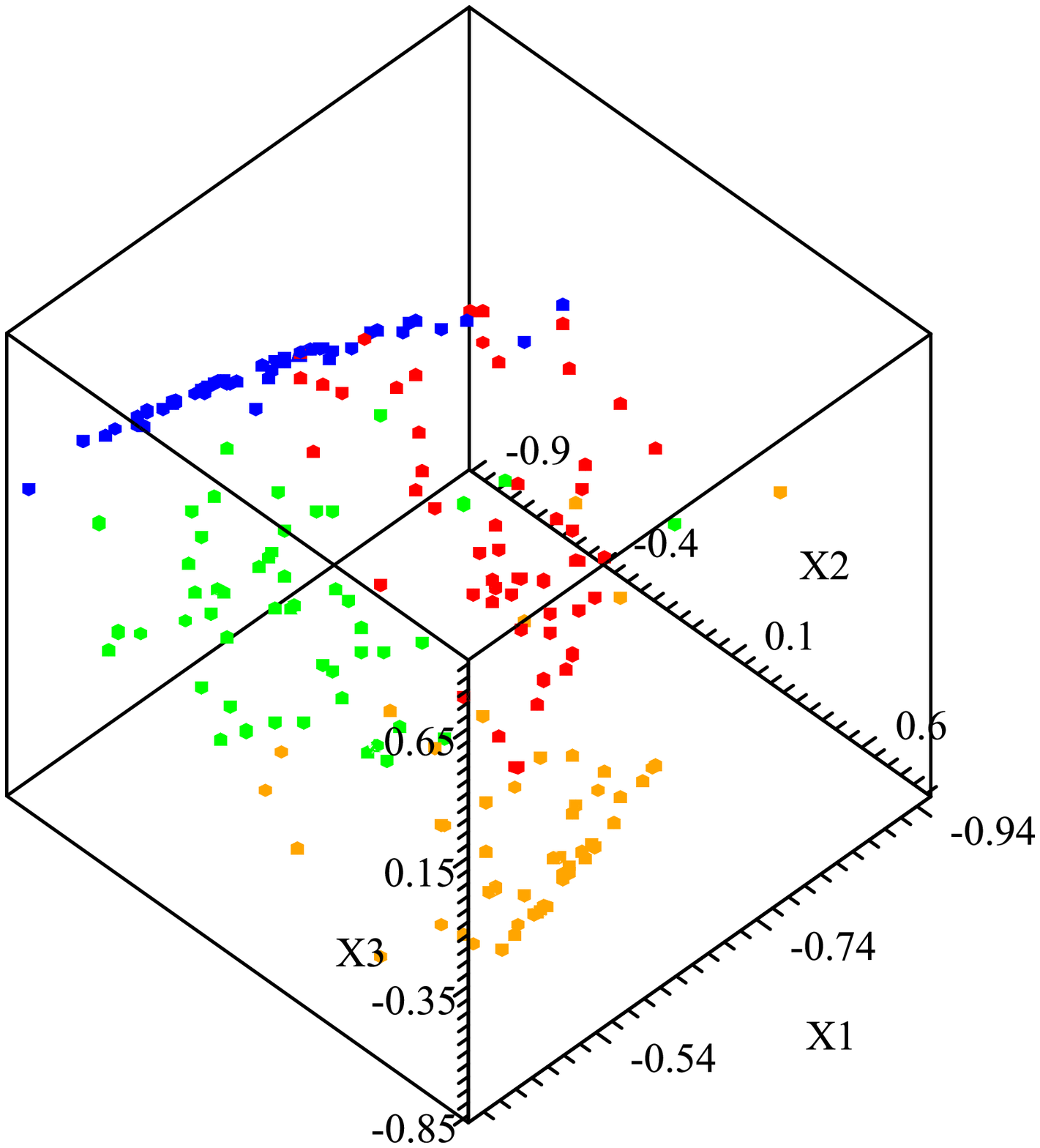}
  \hss\qquad\hss
  \includegraphics[width=1.5in]{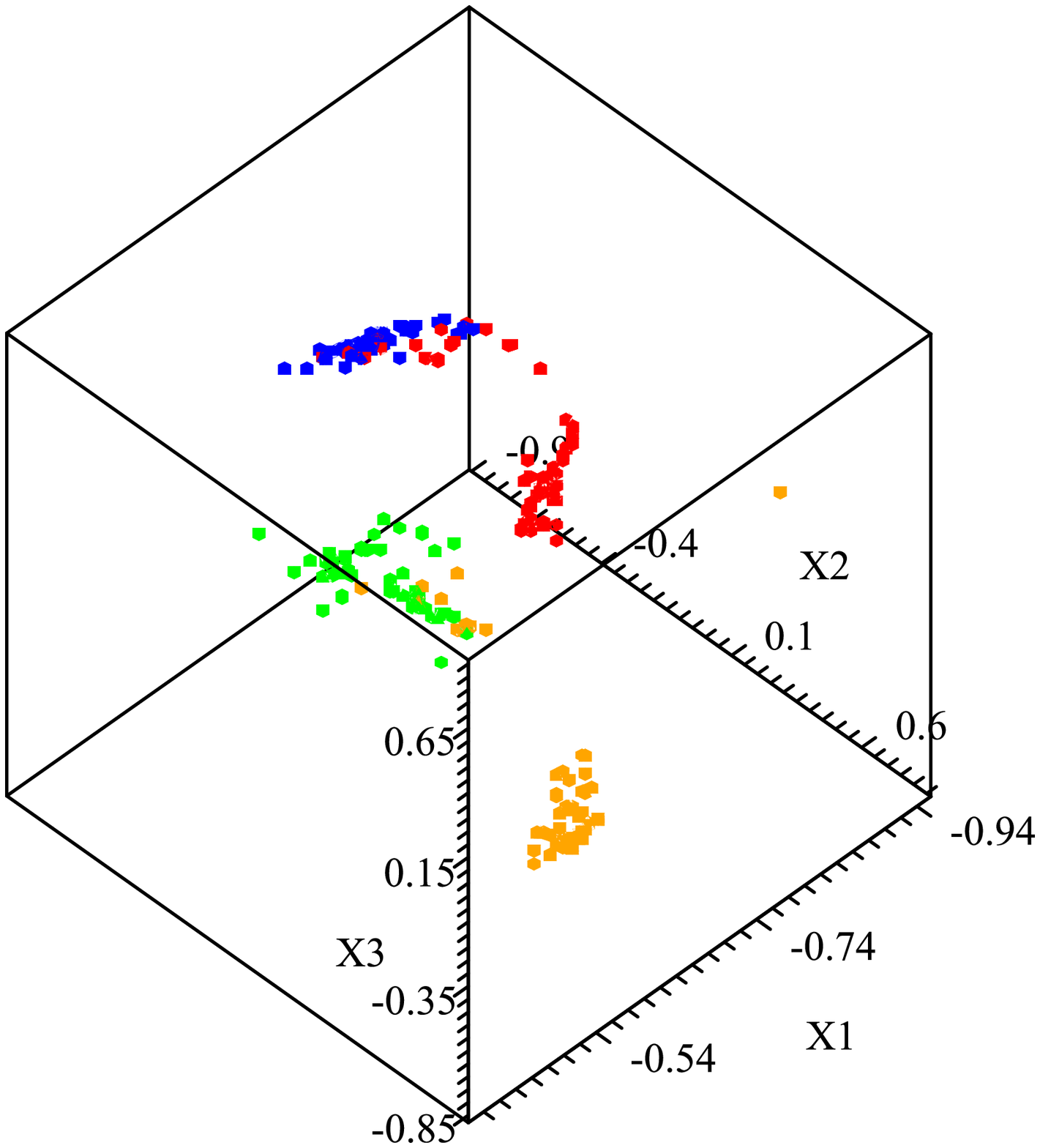}
   \hss}\hbox to \hsize{\hss
   \includegraphics[width=2.0in]{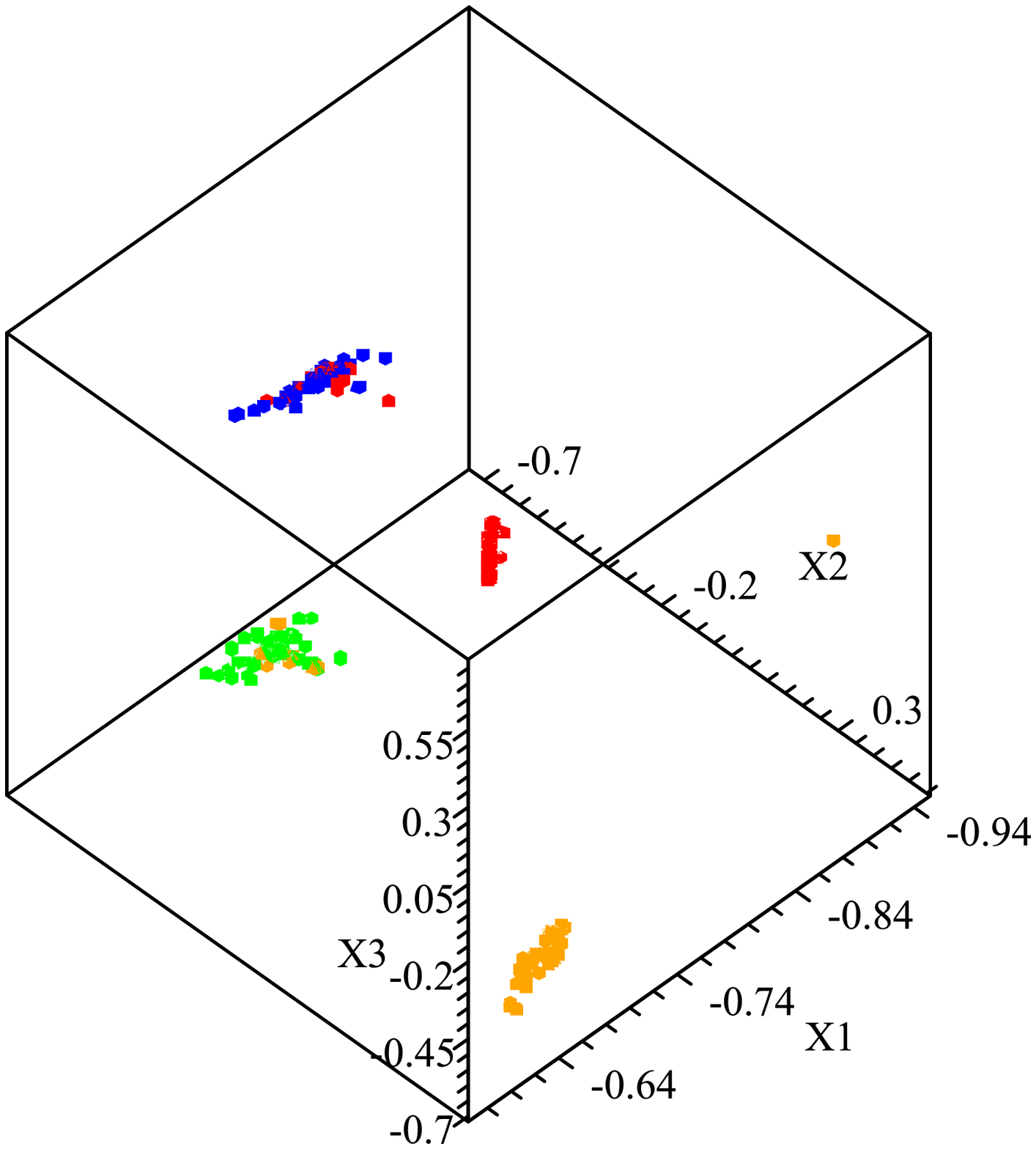}
    \hss}
  }
  \caption{The left hand plot shows three-dimensional distribution of the
  original data points before quantum evolution.  The middle plot shows
  the same distribution after quantum evolution. The right hand plot shows the results
of an additional iteration of DQC. The values of parameters
 used to construct the Hamiltonian and evolution operator are:
  $\sigma = 0.07$ and $m = 0.2$. Colors indicate the
  expert classification of data into four classes, unknown to the clustering algorithm.
  Note, small modifications of the parameters lead to the same results.}
\label{new1}
\end{figure}

\section{Exploiting DQC}

Obviously, in this brief overview of DQC, I have not given you all
the details of how to implement the algorithm, but I have given you
all of the essentials.  Now I want to point out that although colors
in Fig.\ref{new1} were there so that you could see the clusters form
and tell how well DQC was working, the fact that colors can be used
in the visualization step can play an additional role in data
mining.  The crab problem was an example of a blind search for
clusters. What I want to concentrate on now is a different sort of
problem; namely, a situation where we start with a classification of
the data into clusters, but we do not know how the measured features
in the data-set relate to the classification.  As an example,
consider Affymetrix gene chip data for a set of Leukemia patients.
The Affymetrix gene chip is a silicon device to which strands of
RNA are attached in a regular matrix. The setup is made so that,
if a protein binds to a strand of RNA that codes for that protein
the the spot to which the strand is bound flouresces when light
shines on the chip.  Such a gene chip can measure
the expression of over 7000 genes (by seeing the proteins that
are coded for by that gene).  Now, in the case of ALL and AML
leukemia cells we start with a clinical classification based upon
a pathological examination of the cells.  So we know how to color
spots in the DQC picture according to type.  What we don't know
is how to use the gene chip information to identify these cells;
that is what we want to do using DQC (or any other clustering
method).  Two problems that one has, in addition to the clustering
problem, is that the binding of a protein to a strand of
RNA is not all that specific, and most of the genes being
measured have nothing to do with cancer.  That is where the
coloring comes in.  If one applies DQC to such a data set and
sees clusters of the correct color form, then one knows that
the gene chip data contains the information we need.  The next
step is to eliminate features (i.e., measurements of particular
genes) without hurting the clustering.  (There are various schemes
for doing this, including the brute force approach.)  Clearly,
if we pursue this process and weed out genes that have nothing to
do with the clustering, then one goes a long way to both developing
a diagnostic tool, and obtaining an insight into which genes are
related to the specific cancer.

I will show you some results for a data set by Golub {\it et.al.}\cite{golub}.
This set contains gene chip measurements on cells from 72 leukemia
patients with two different types of Leukemia, ALL and AML. The
expert identification of the classes in this data set is based upon
dividing the ALL set into two subsets corresponding to T-cell and
B-cell Leukemia.  The AML set is divided into patients who underwent
treatment and those who did not.  In total the Affymetrix GeneChip
used in this experiment measured the expression of 7129 genes. The
specific feature filtering method we employed in this analysis was
based on SVD-entropy, and is a simple modification of a method
introduced by Varshavsky {\it et al.}\cite{featsel} and applied to
the same data.  For our purposes it doesn't matter what the feature
filtering method was, I want you to see the difference removing features
makes in clustering.  I also want you to recognize how the DQC
visualization makes it easy to quickly identify the effect of
removing features from the data.
\begin{figure}[h!]
   \hbox to \hsize{\hss
   \includegraphics[width=1.5in]{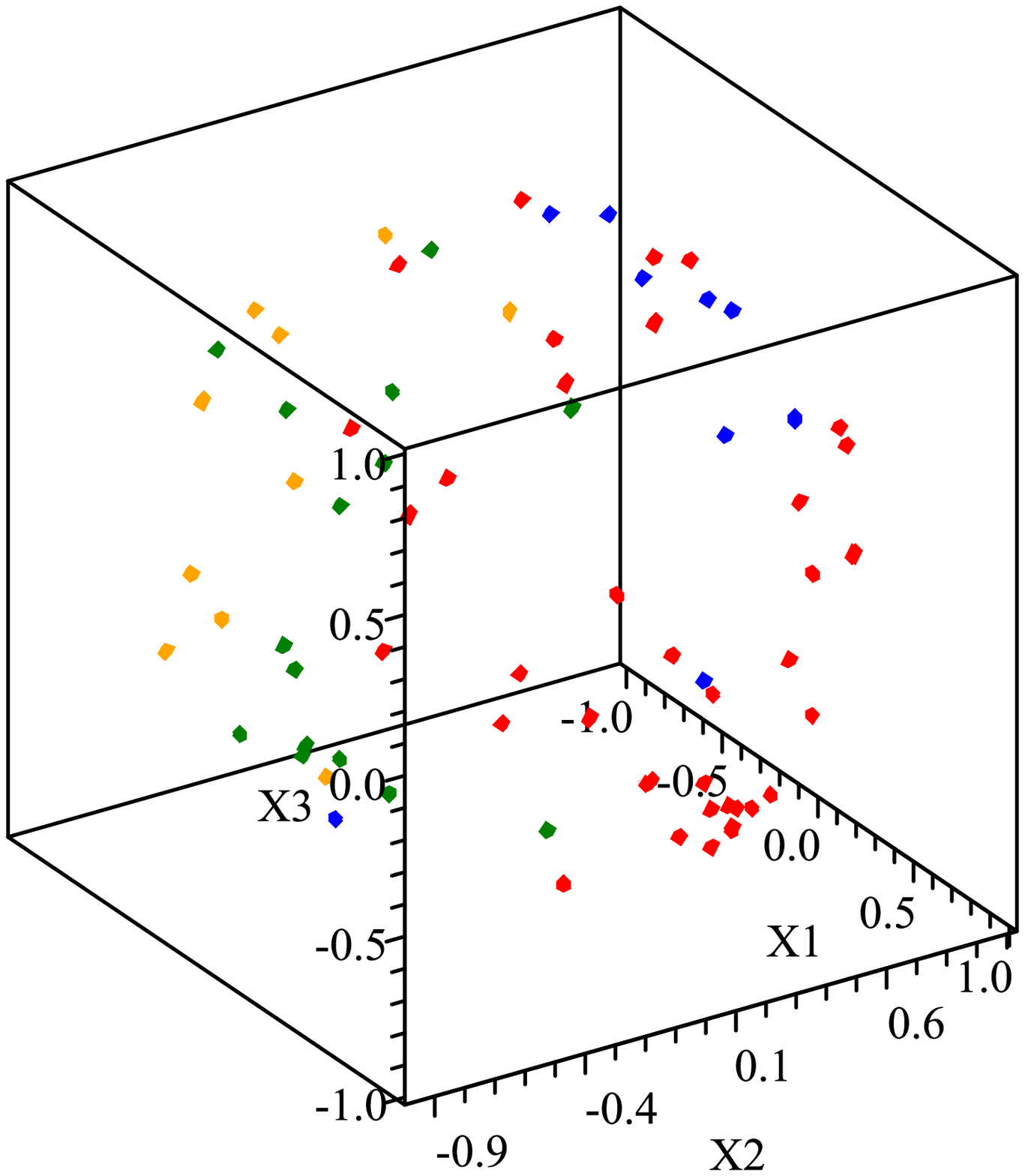}
  \hss\quad\hss
   \includegraphics[width=1.5in]{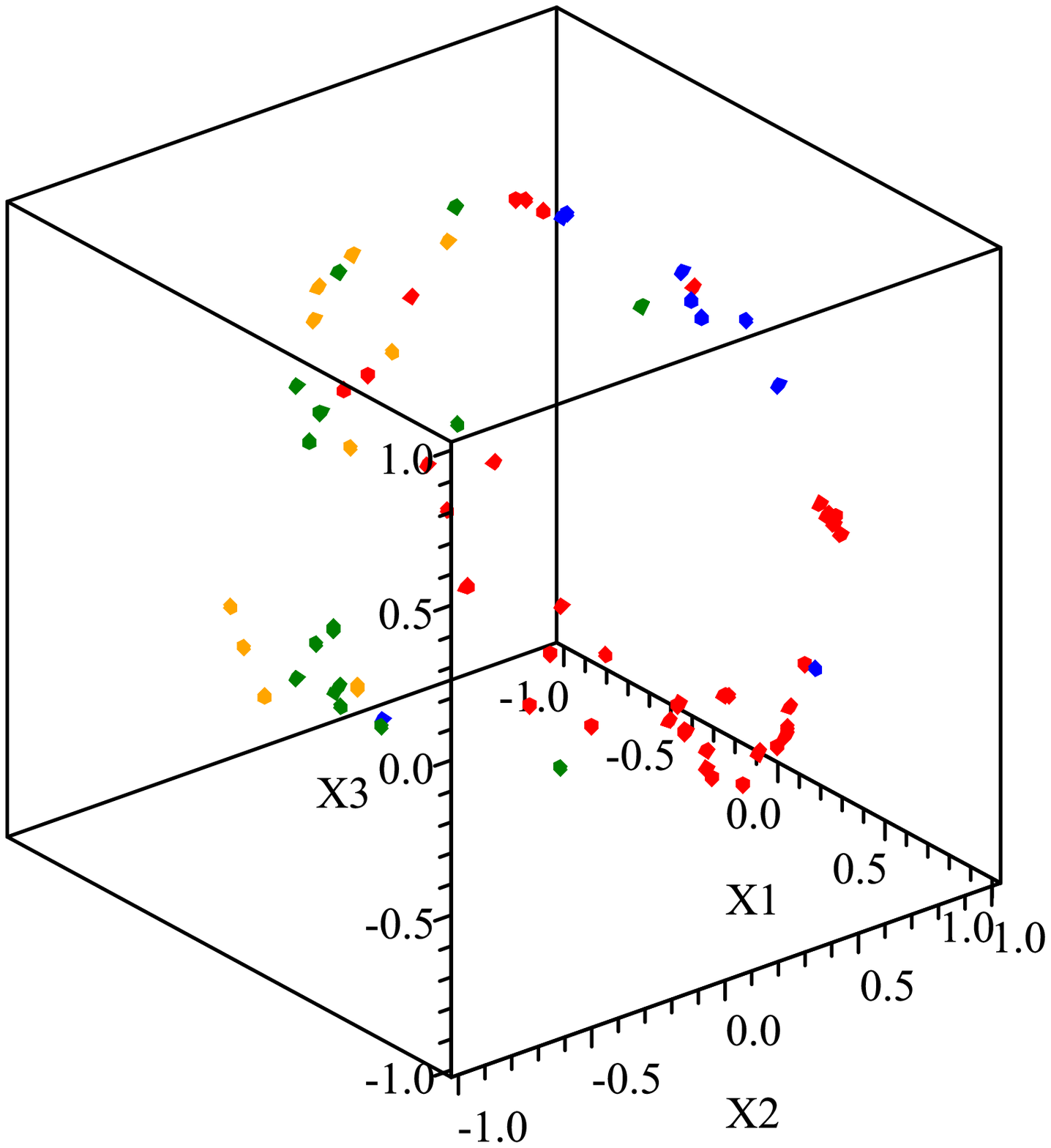}
    \hss}
  \caption{The left hand picture is the raw data from the
  Affymetrix Chip plotted for principal components 2,3,4.
  Clearly, without the coloring it would be hard to identify
  clusters.  The right hand picture is the same data after DQC
  evolution using $\sigma =0.2$ and a mass $m=0.01$.  The different
  classes are shown as blue, red, green and orange.
   }
\label{GolubOne}
\end{figure}

\begin{figure}[h!]
   \hbox to \hsize{\hss
   \includegraphics[width=1.5in]{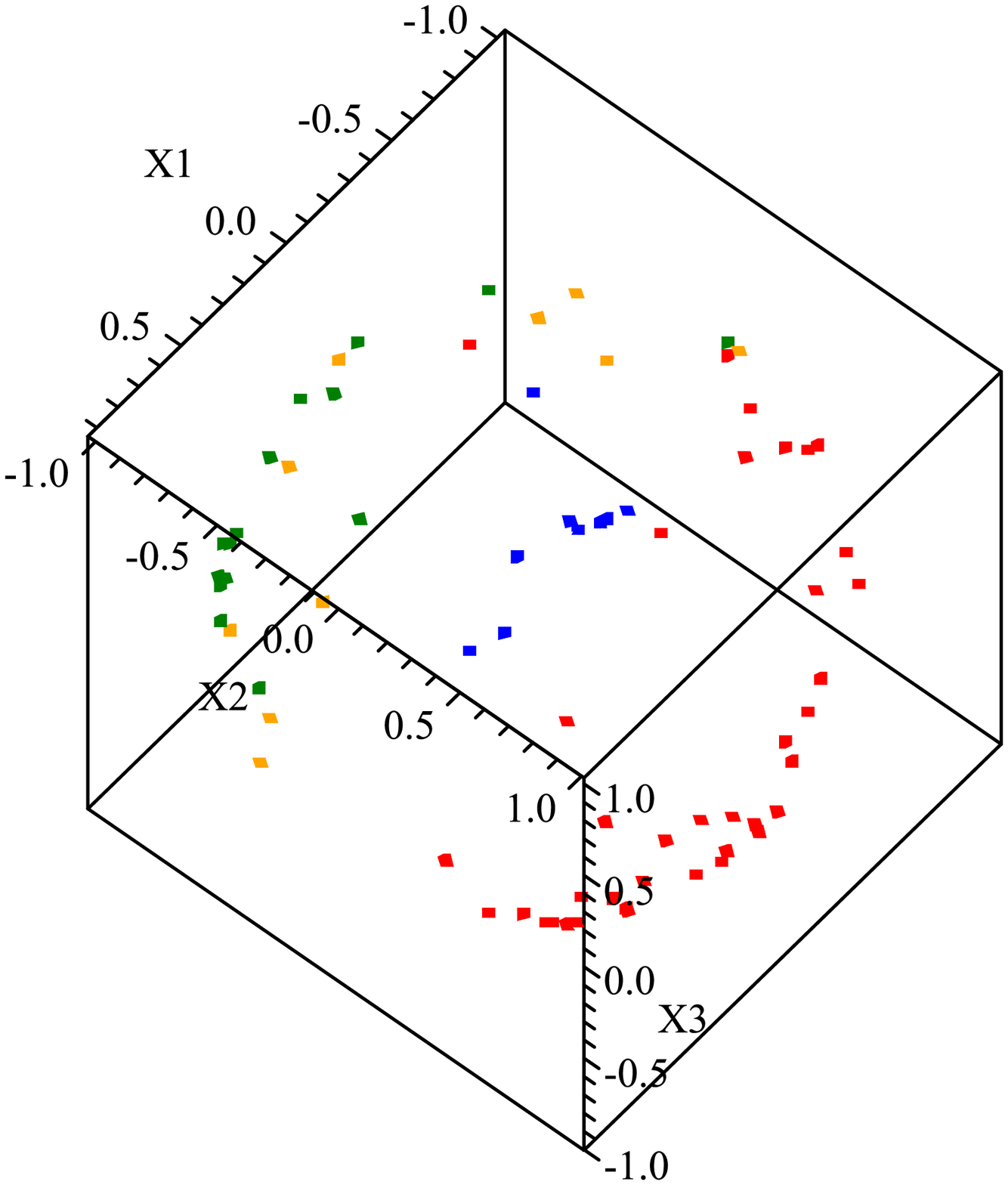}
  \hss\quad\hss
   \includegraphics[width=1.5in]{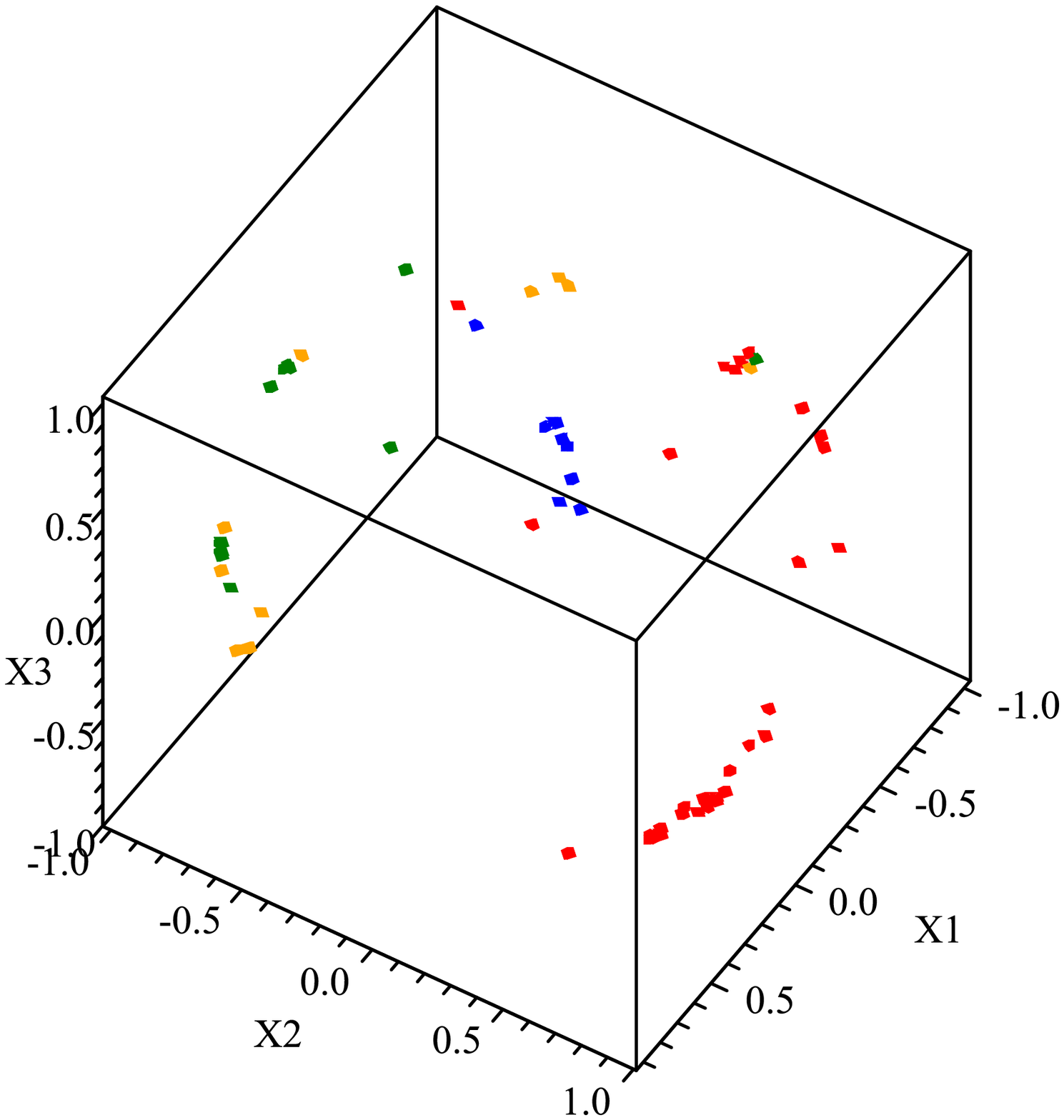}
    \hss}
  \caption{The left hand plot is the Golub data after one stage
  of SVD-entropy based filtering, but before DQC evolution.  The right
  hand plot is the same data after DQC evolution.
   }
\label{GolubTwo}
\end{figure}
Figure~\ref{GolubOne} displays the raw data in the 3-dimensional
space defined by the second to fourth principal components,
and the effect that DQC has on these
data. In Figure~\ref{GolubTwo} we see the result of applying feature
filtering to the original data, represented in the same
3-dimensions, followed by DQC evolution.  Applying a single stage of
filtering has a dramatic effect upon clustering, even before DQC
evolution. The latter helps sharpening the cluster separation.
\begin{figure}[h]
   \hbox to \hsize{\hss
   \includegraphics[width=1.5in]{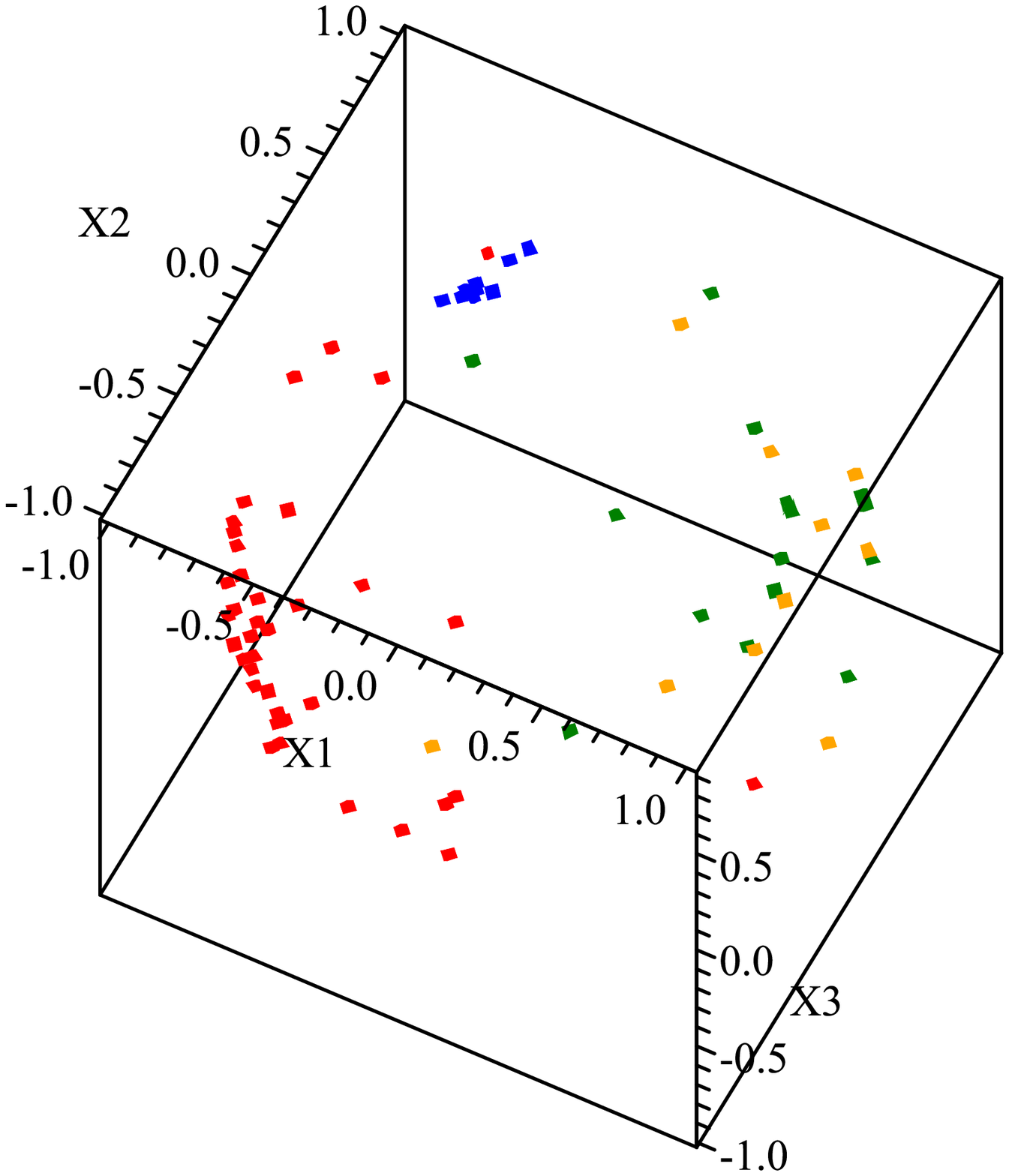}
  \hss\quad\hss
   \includegraphics[width=1.5in]{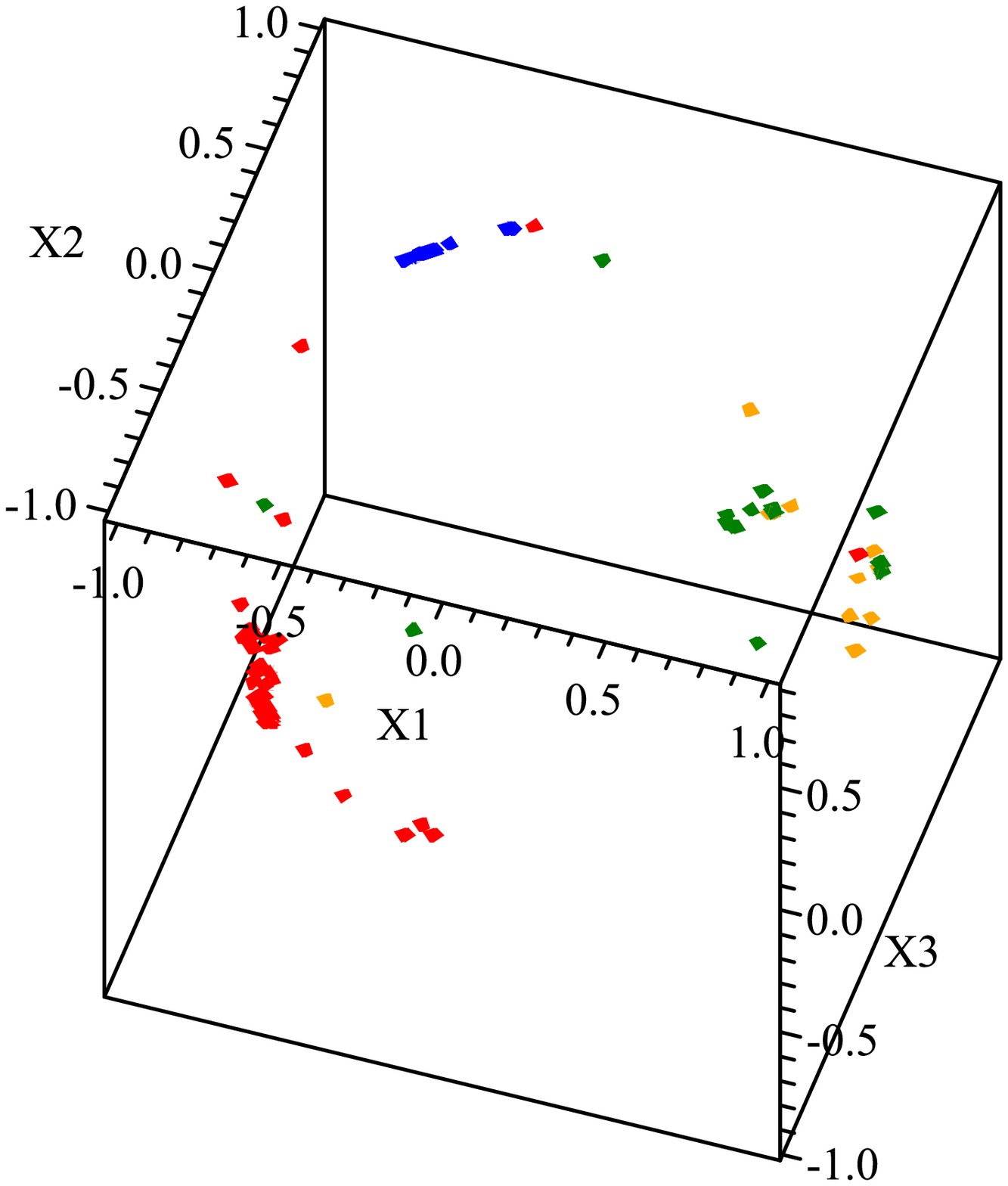}
    \hss}
  \caption{The left hand plot is the data after three stages
  of SVD-entropy based filtering, but before DQC evolution.  The right
  hand plot is the same data after DQC evolution.
   }
\label{GolubThree}
\end{figure}
Figure \ref{GolubThree} shows the results of removing many features
(using the SVD entropy method) before and after DQC evolution. These
plots, especially the after DQC pictures, show dramatic clustering,
especially for the blue points.  With each stage of filtering we see
that the blue points cluster better and better, in that the single
red outlier separates from the cluster and the cluster separates
more and more from the other points.  The blue points are what I
refer to as an {\it obviously robust cluster\/} identified in early
stages of feature filtering. If one continues removing features,
however, the clear separation of the blue points from the others
begins to diminish. This tells us that we have gone to far with the
blue points, we are now removing features that matter for its
classification. This is, of course, just what we are looking for, a
way of identifying those features which are important to the
existing biological clustering. We could, at this juncture, search
among the most recent 278 eliminated features to isolate those most
responsible for the separation of the blue cluster from the others.
Now however I want to make another point.  Since the blue cluster is
so robust and easily identified, let us remove the blue cluster from the
original data and repeat the same process without this cluster.  The
idea here is that now the SVD-entropy based filtering will not be
pulled by the blue cluster and so it will do a better job of sorting
out the red, green and orange clusters.  I want you to see that this is in
fact the case.
\begin{figure}[h]
   \hbox to \hsize{\hss
   \includegraphics[width=1.5in]{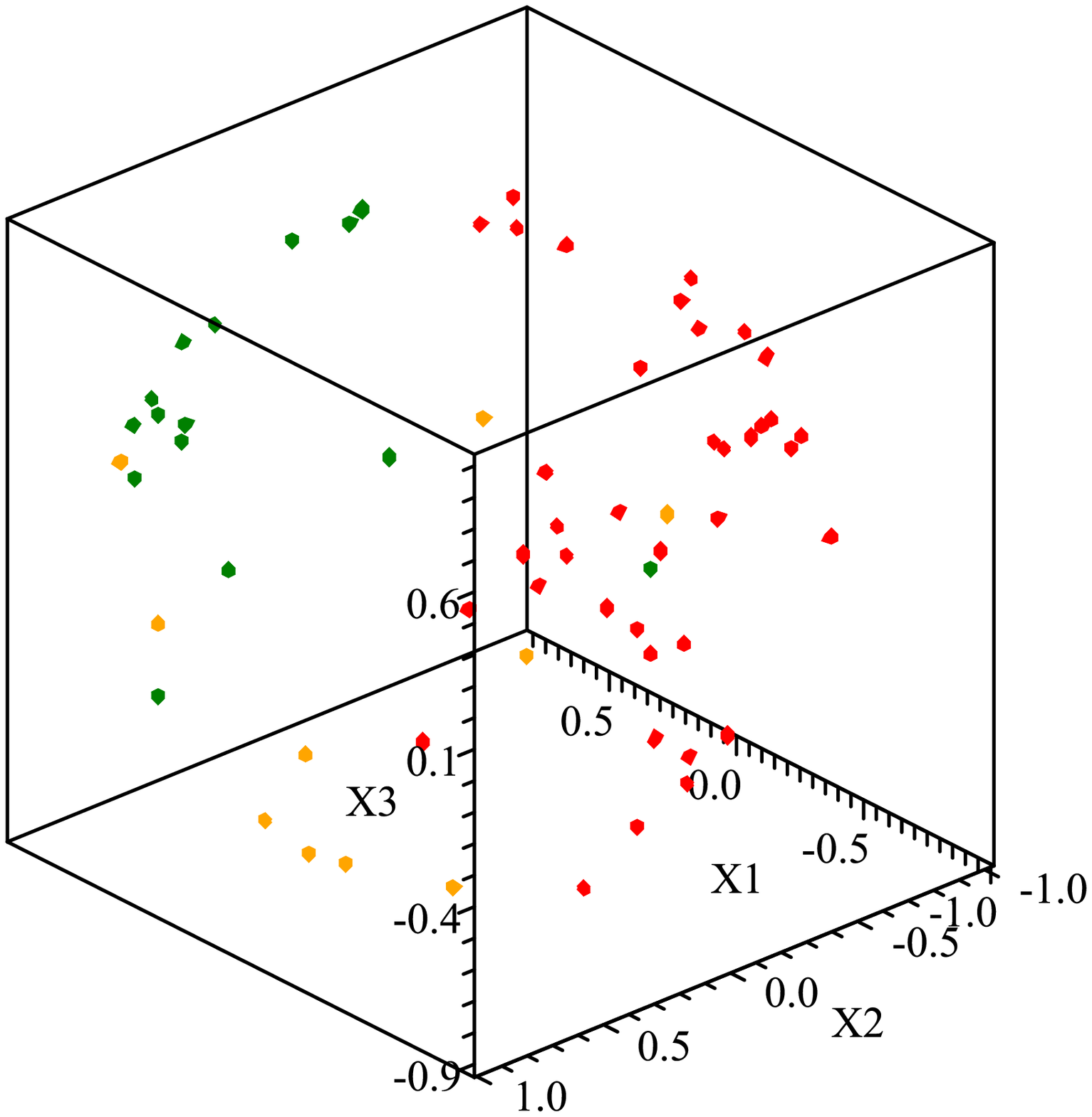}
  \hss\quad\hss
   \includegraphics[width=1.5in]{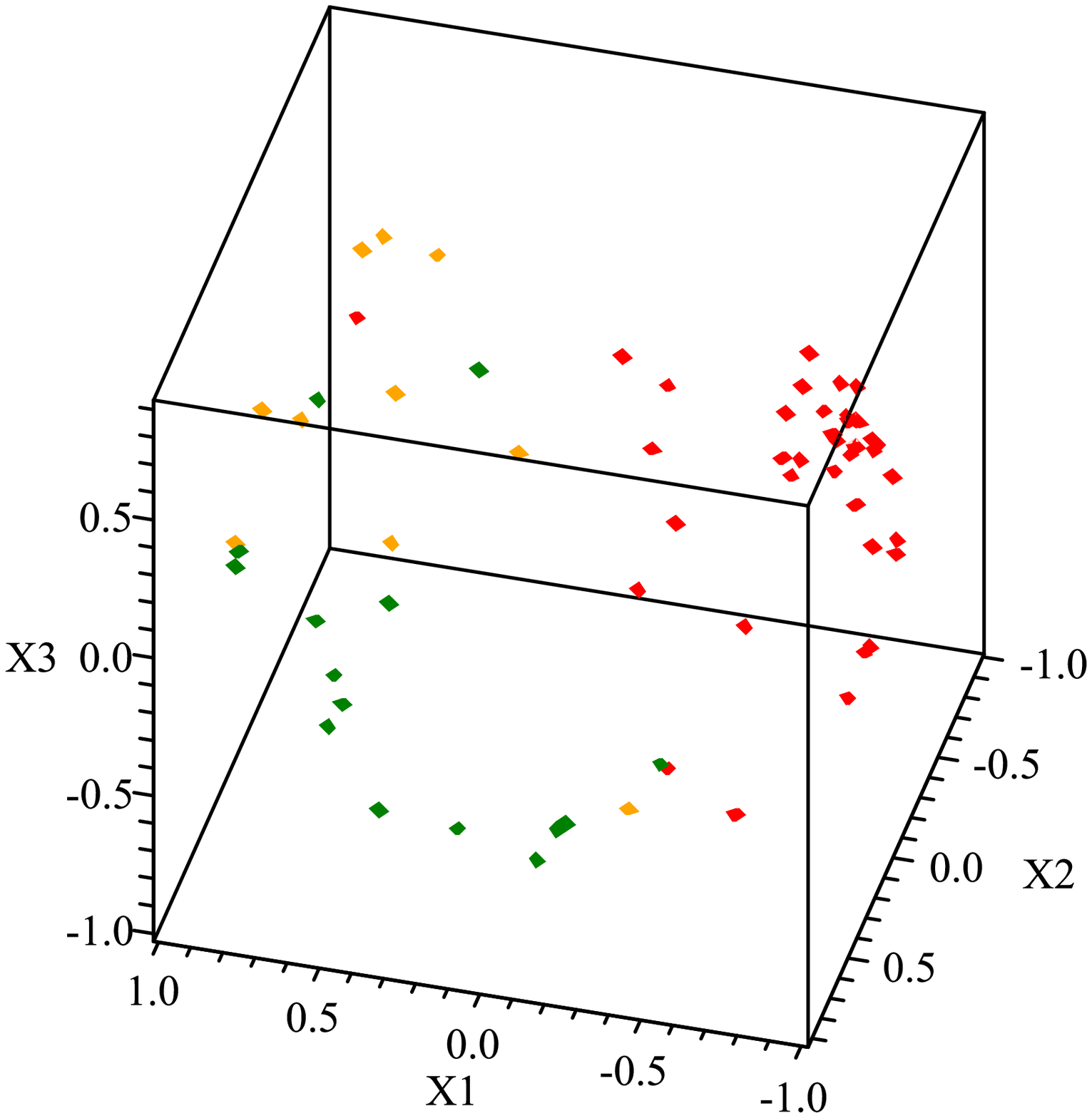}
    \hss}
  \caption{The left hand plot is what the starting data looks like
  if one first removes the blue points and does one stage of SVD-entropy
  based filtering.  The right hand plot is what the starting data looks
  like after three stages of filtering.
   }
\label{GolubFive}
\end{figure}
In Figure \ref{GolubFive} we see a plot of what the starting
configurations look like if one takes the original data, removes the
blue cluster and re-sorts the reduced data set according
to the SVD-entropy based filtering rules.  The left hand plot is
what happens if one filters a single time. The right hand plot
shows what happens if one repeats the filtering procedure two more times,
It is clear from the plots that each iteration of the
filtering step improves the separation of the
starting clusters.  By the time we have done five filtering steps the
red, green and orange clusters are distinct, if not obviously separated.
\begin{figure}
   \hbox to \hsize{\hss
   \includegraphics[width=1.5in]{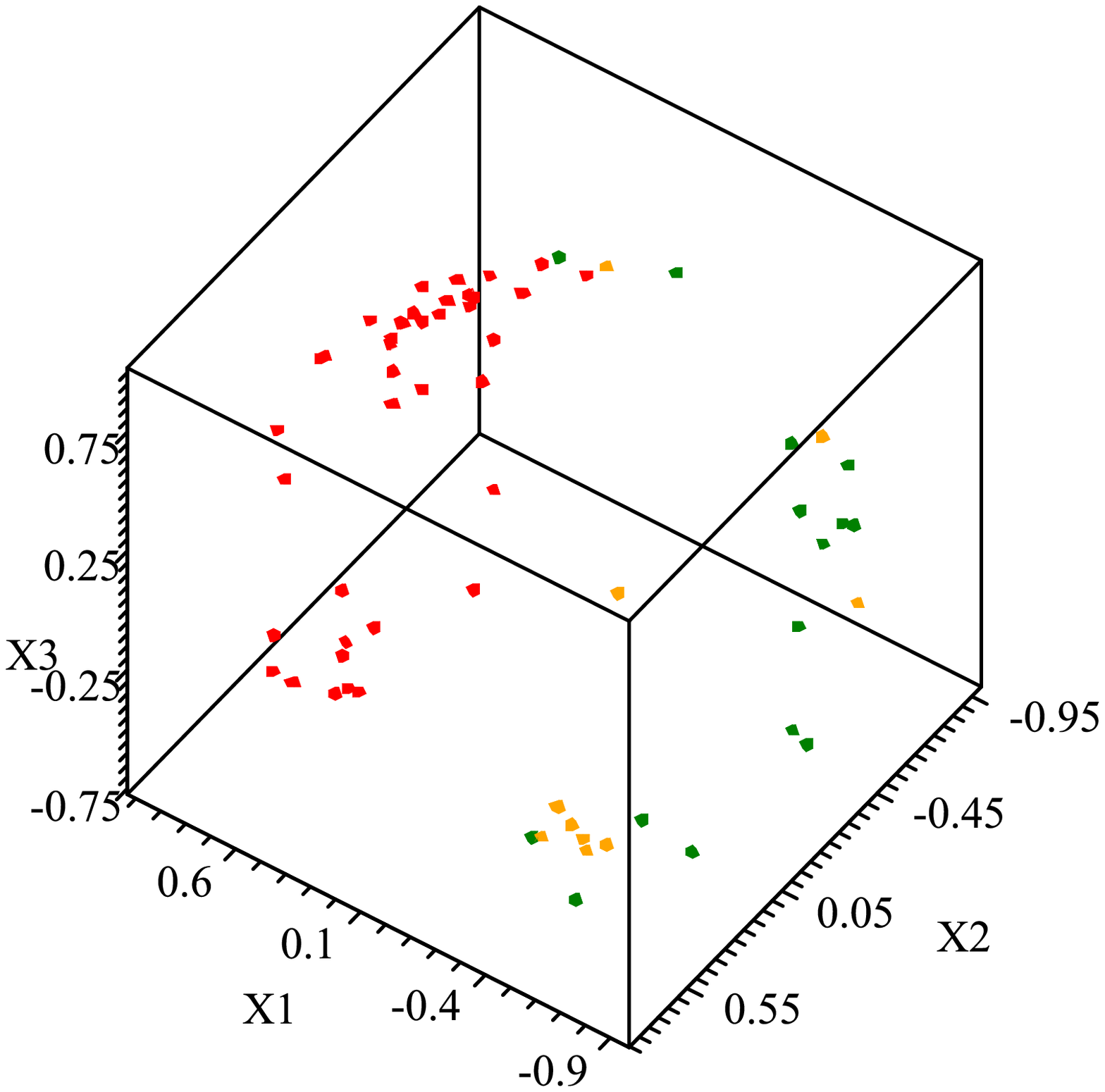}
  \hss\quad\hss
   \includegraphics[width=1.5in]{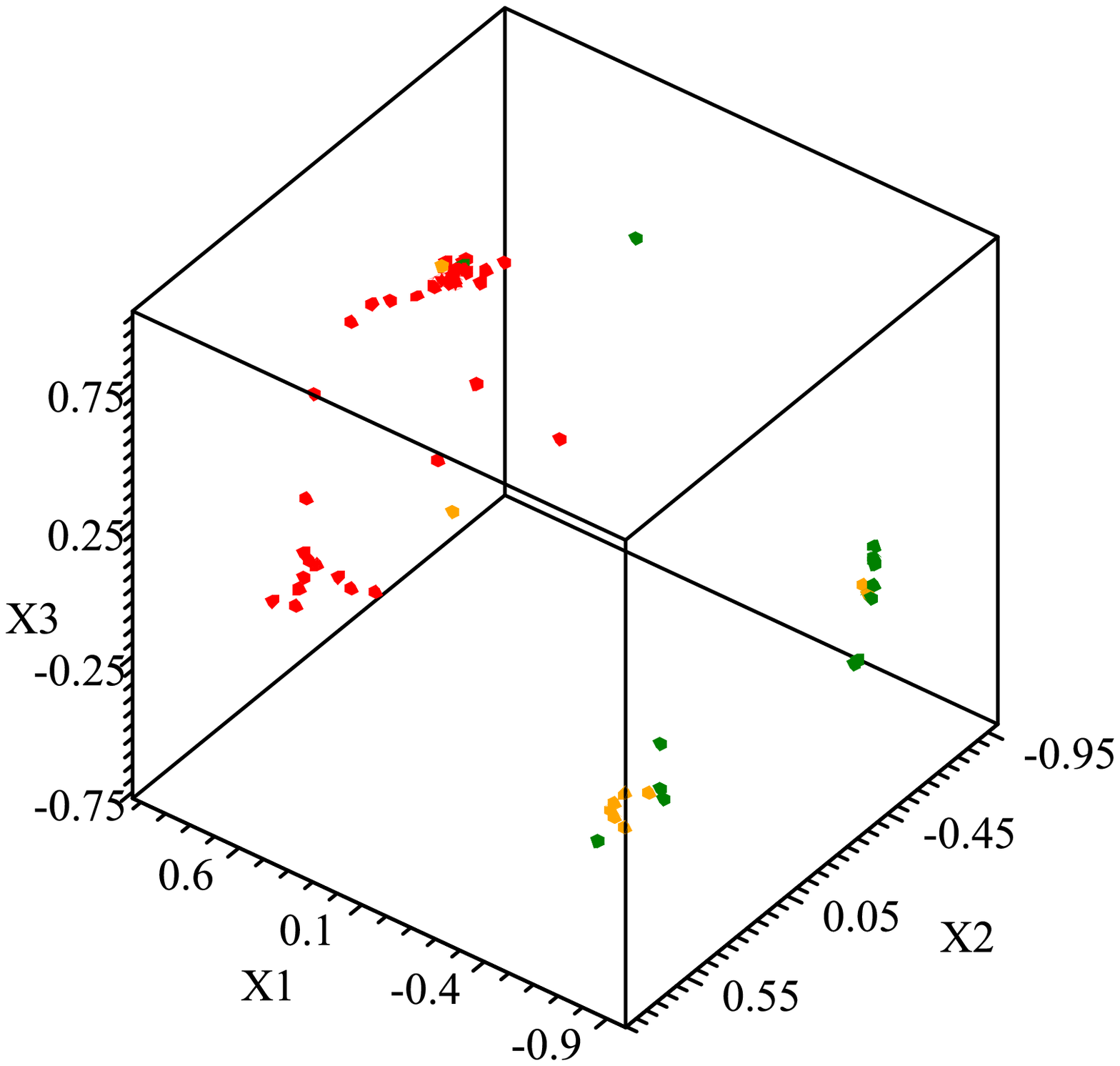}
    \hss}
   \hbox to \hsize{\hss
   \includegraphics[width=1.5in]{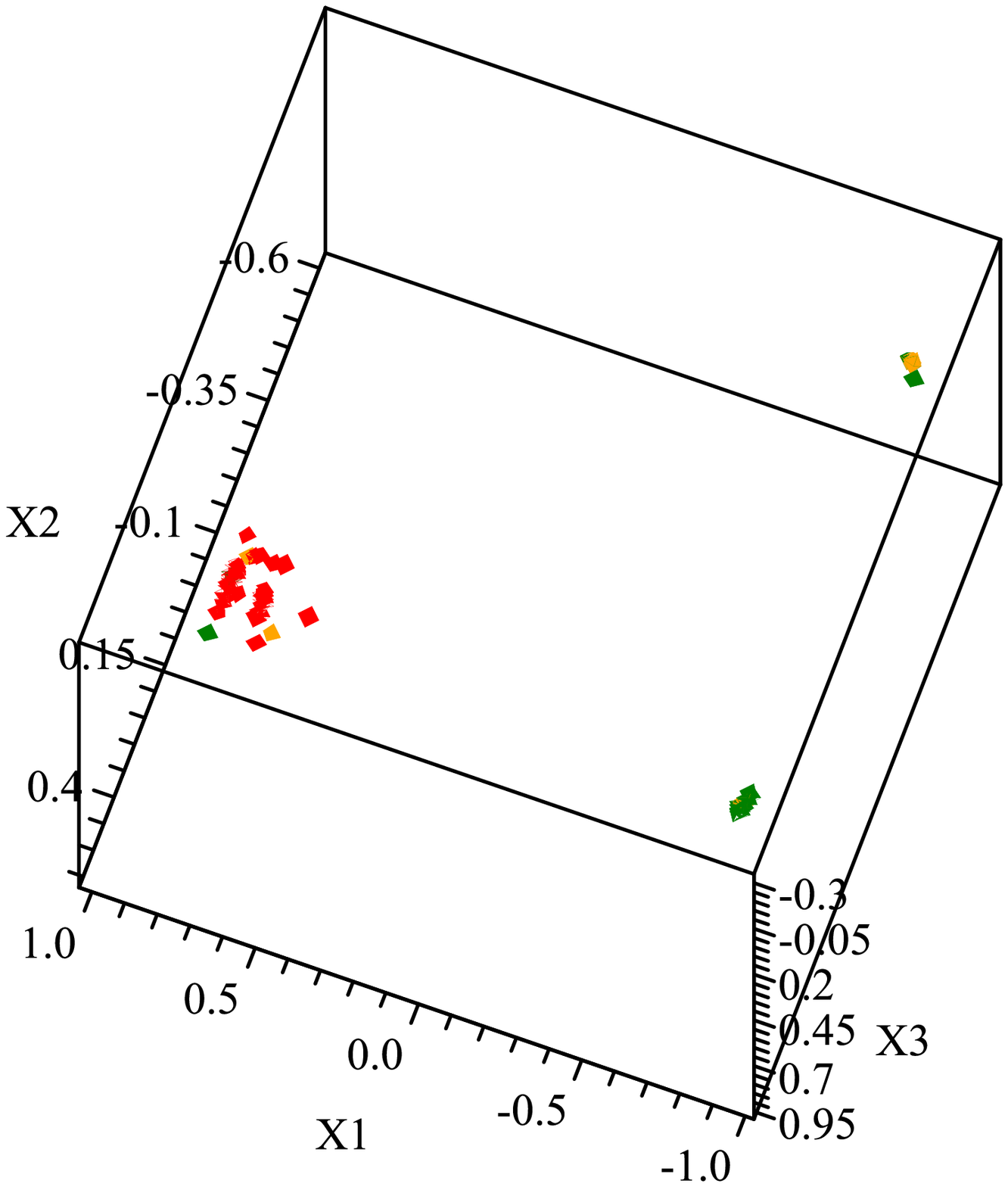}
   \hss}
  \caption{The left hand plot is what the starting data looks like
  if one first removes the blue points and does five stages of SVD-entropy
  based filtering.  The right hand plot is what happens after one stage
  of DQC evolution.  The bottom plot is the final result after iterating
  the DQC evolution step two more times.  At this point the clusters
  are trivially extracted.
   }
\label{GolubSix}
\end{figure}
Finally, to complete our discussion, we show Figure \ref{GolubSix}.
This figure shows the results of doing five iterations of the
SVD-entropy based filtering and following that with three stages of
DQC evolution.  The dramatic clustering accomplished by DQC
evolution makes it easy to extract clusters.  Note however, that in
the second plot we see what we have seen throughout, that the red
points first form two distinct sub-clusters which only merge after
two more stages of DQC evolution.  This constant repetition of the
same phenomenon, which is only made more apparent by SVD-entropy based
filtering, is certainly a real feature of the data.  It presumably
says that what appears to be a sample of a single type of cell at
the biological level is in reality two somewhat different types of
cells when one looks at gene expression. A measure of the success of clustering
is given by the Jaccard score which, for this result
is $0.762$, higher than the value 0.707 obtained by \cite{featsel}.

\section{DQC and Large Data Sets}

There are many scientific and commercial fields, such as cosmology,
epidemiology, particle physics, risk-management, etc., where the
one deals with very large data sets, often in large
numbers of dimensions. DQC, by its nature, doesn't have
trouble with large dimensions.  In general it allows one to use
SVD decomposition with much less severe dimensional reductions
than other methods.  However, dealing with large number of points
requires some thought.

Since the method for evolving point requires diagonalizing the
truncated Hamiltonian, and since diagonalizing such a matrix
on a PC becomes difficult for matrices that are larger than
$2000 \times 2000$, it is obvious that using brute force
methods to evolve sets of data having tens or hundreds of
thousands of points simply won't work.  The solution
to this problem lies in the fact that
the SVD decomposition maps the data into an $N$-dimensional
unit cube, and the fact that the data points are represented by states in
Hilbert space rather than $N$-tuples of real numbers.

The key observation is that, since Gaussian wavefunctions whose
centers lie within a given cube have non-vanishing overlaps, as one
chooses more and more Gaussians one eventually arrives at a
situation where the states become what we will refer to as {\it
essentially linearly dependent\/}.  In other words, we arrive at a
stage at which any new wave-function added to the set can, to some
predetermined accuracy, be expressed as a linear combination of the
wave-functions we already have.  Of course, since quantum mechanical
time evolution is a linear process, this means that these additional
states can be evolved by expressing them as linear combinations of the
previously selected states and using the evolution of those states
to evolve the extra states.  Since computing the overlap of two
Gaussians is done analytically determining which points determine
the set of {\it maximally essentially linearly independent states
for the problem\/} is easy. Typically, even for data sets with
$100,000$ points, this is of the order of $1600$ points.  The small
number works because, as we have already noted, we don't need high
accuracy for DQC evolution. The quality of the clustering degrades
very slowly with loss in accuracy. Thus, it is possible to compute the time
evolution operator in terms of a well chosen subset of the data and
then apply it to the whole set of points.

\section{Afterward}

Well this is about all I can cover in my allotted time.  I hope
I have convinced you that, strange as it may seem, quantum mechanics
and data mining are related to one another.  In fact, there is
enough interest in this stuff in the real world that Stanford(SLAC)
and Tel-Aviv University are patenting this technology and some
companies have already expressed an interest in using it.

\end{document}